\documentclass[10pt,twocolumn,letterpaper]{article}

\usepackage{wacv}
\usepackage{times}
\usepackage{epsfig}
\usepackage{graphicx}
\usepackage{amsmath}
\usepackage{amssymb}
\usepackage[dvipsnames]{xcolor}
\usepackage{array}
\usepackage{multirow}
\usepackage{siunitx}
\usepackage[percent]{overpic}
\usepackage{fdsymbol}

\newcolumntype{P}[1]{>{\centering\arraybackslash}p{#1}}
\newcommand\tstrut{\rule{0pt}{2.4ex}}

\makeatletter
\newlength\mylen
\DeclareRobustCommand*\mytextsuperscript[1]{%
	\@Textsuperscript{\selectfont#1}}
\def\@Textsuperscript#1{%
	\settoheight\mylen{\fontsize\f@size\z@ A}%
	{\m@th\ensuremath{\raise.5\mylen\hbox{\fontsize\sf@size\z@#1}}}}
\makeatother

\usepackage[pagebackref=true,breaklinks=true,letterpaper=true,colorlinks,bookmarks=false]{hyperref}
\wacvfinalcopy 

\def\wacvPaperID{0038} 

\ifwacvfinal\fi
\begin{document}

\title{SVIRO: Synthetic Vehicle Interior Rear Seat Occupancy \\ Dataset and Benchmark}

\author{Steve Dias Da Cruz\,\mytextsuperscript{1,2,3}\\
	{\tt\small steve.dias-da-cruz@iee.lu}
	\and
	Oliver Wasenm\"uller\,\mytextsuperscript{2}\\
	{\tt\small oliver.wasenmueller@dfki.de}
	\and
	Hans-Peter Beise\,\mytextsuperscript{4}\\
	{\tt\small h.beise@inf.hochschule-trier.de}
	\and
	Thomas Stifter\,\mytextsuperscript{1}\\
	{\tt\small thomas.stifter@iee.lu}
	\and
	Didier Stricker\,\mytextsuperscript{2,3}\\
	{\tt\small didier.stricker@dfki.de} \\
	\and
	\mytextsuperscript{1}\,IEE S.A. \hspace{0.5cm} \mytextsuperscript{2}\,DFKI - German Research Center for Artificial Intelligence \\
	\mytextsuperscript{3}\,University of Kaiserslautern \hspace{0.5cm} \mytextsuperscript{4}\,Trier University of Applied Sciences 
}
\maketitle
\ifwacvfinal\thispagestyle{empty}\fi

\begin{abstract}
We release SVIRO, a synthetic dataset for sceneries in the passenger compartment of ten different vehicles, in order to analyze machine learning-based approaches for their generalization capacities and reliability when trained on a limited number of variations (e.g. identical backgrounds and textures, few instances per class). This is in contrast to the intrinsically high variability of common benchmark datasets, which focus on improving the state-of-the-art of general tasks. Our dataset contains bounding boxes for object detection, instance segmentation masks, keypoints for pose estimation and depth images for each synthetic scenery as well as images for each individual seat for classification. The advantage of our use-case is twofold: The proximity to a realistic application to benchmark new approaches under novel circumstances while reducing the complexity to a more tractable environment, such that applications and theoretical questions can be tested on a more challenging dataset as toy problems. The data and evaluation server are available under \url{https://sviro.kl.dfki.de}.
\end{abstract}
\section{Introduction}
\label{Section:intro}
\begin{figure}
	\begin{center}
		\begin{overpic}[width=0.495\linewidth]{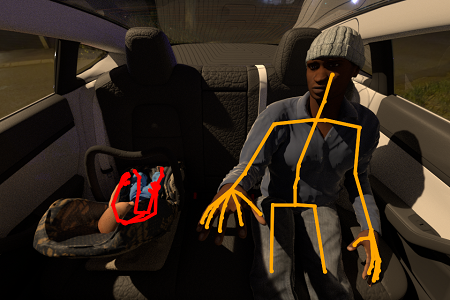}
			\put(4,4.7){\Large\textcolor{white}{a}}
		\end{overpic}
		\begin{overpic}[width=0.495\linewidth]{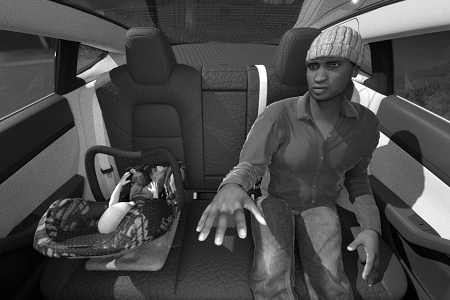}
			\put(4,4.7){\Large\textcolor{white}{b}}
		\end{overpic}
		\vskip 0.2 \baselineskip
		\begin{overpic}[width=0.495\linewidth]{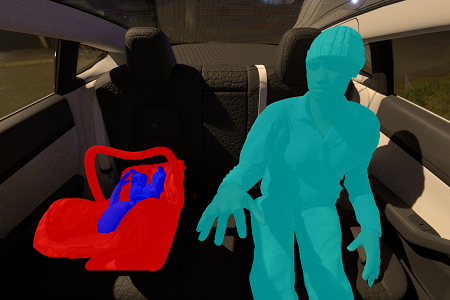}
			\put(4,4.7){\Large\textcolor{white}{c}}
		\end{overpic}
		\begin{overpic}[width=0.495\linewidth]{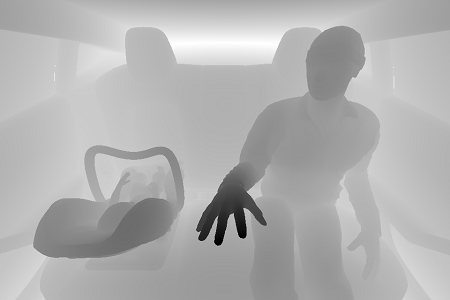}
			\put(4,4.7){\Large\textcolor{white}{d}}
		\end{overpic}
	\end{center}
	\caption{Example scenery of SVIRO together with the provided ground truth data. Left seat: infant seat with an infant. Middle seat: empty. Right seat: adult passenger. a) RGB image with keypoints for human pose estimation. b) Grayscale infrared imitation. c) Position and class based instance segmentation. d) Depth map.}
	\label{fig:example_real_orss}
\end{figure}
Interior vehicle sensing has gained increased attention in the research community, in particular due to challenges and developments related to automated vehicles \cite{7501845, fridman2017autonomous}. In this work, we focus on rear seat occupant detection and classification using a camera system and different ground truth data, as illustrated in Figure \ref{fig:example_real_orss}. Information about the presence and location of the passengers can be used to help reduce injuries in case of an accident, e.g. by adjusting the strength of airbag deployment \cite{airbag, perrett2016cost}. Seat occupancy detection can be used to remind the passengers to fasten their seat-belts or to detect children forgotten in the car \cite{da2019theoretical, diewald2016rf}. For autonomous driving, it will be of interest to understand the overall scenery in the car interior \cite{pulgarin2017drivers}, e.g. for handover situations \cite{mccall2016towards}. For all the aforementioned applications, one has to ensure that trained machine learning models will be capable of classifying new types of child seats correctly while not being mislead by arbitrary everyday objects or through the window background sceneries. However, machine learning-based models, and specifically neural networks, trained in a single environment take non-relevant characteristics of the specific environmental conditions into account in an uncontrolled way \cite{tian2018eliminating} and therefore data must be recorded repetitively for different environments. Acquiring images in various (natural) lightning and weather conditions and accounting for different seat textures, car interior features, or even changing camera poses make the data acquisition even more difficult. While domain adaptation investigates solutions to account for a shift in the source distribution with respect to the target distribution, common approaches still need a large amount of data for the target distribution \cite{sun2016deep, tzeng2017adversarial} to work well. Consequently, the means for generating a real training dataset with the corresponding annotations are limited and need to be repeated for each additional new car model and automotive manufacturer. Therefore, theoretically founded means to overcome the limitations of datasets collected for many real world applications have to be developed or advanced.

Common machine learning datasets and benchmarks focus on pushing the state-of-the-art of general tasks like classification \cite{deng2009imagenet}, segmentation \cite{Cordts2016Cityscapes}, object detection \cite{braun2019eurocity}, human pose estimation \cite{fabbri2018learning} or multiple tasks at once \cite{Everingham10, Geiger2012CVPR, OpenImages, lin2014microsoft}. They do so on sceneries of high variable backgrounds and intra-class variations, or focus on toy examples to investigate theoretical and fundamental research questions \cite{burgess2019monet}. However, none of the available datasets focuses on the application-oriented case when all images are taken on the same, or similar, background. They do not consider classes with only sparse representations, as is common in engineering problems when the available resources are limited. Consequently, available datasets do not provide a framework to evaluate models trained in the above-mentioned challenging conditions for solving identical tasks, but in a new environment. Hence, similar investigations for the rear seat occupancy cannot be performed and there is no publicly available dataset for the vehicle interior.

We release SVIRO to provide a starting point for investigating the aforementioned challenges and overcome some of the shortcomings of common available datasets. For the training set, we used different human models, child and infant seats, backgrounds and textures than for testing. Hence, we can test the generalization and robustness of models trained in one vehicle to a new one, for solving the same task. Our dataset has a higher visual complexity than toy scenarios while being close enough to a realistic application. Consequently, SVIRO can be used to benchmark common machine learning tasks under new circumstances while allowing the investigation of theoretical questions due to its intrinsically more tractable environment. Additional ground truth data for existing sceneries can be generated or new features can be integrated upon request. For an overview, you can also watch our video \url{https://youtu.be/_arwrYIz7Ok}.

\section{Related work}
\label{Section:relatedwork}

Some previous works have been investigating occupant classification \cite{airbag, perrett2016cost}, seat-belt detection \cite{Baltaxe2019} or skeletal tracking \cite{pulgarin2017drivers} in the passenger compartment, but,  as to best of our knowledge, no dataset was made publicly available. 

Investigations regarding the tasks and challenges as mentioned in Section \ref{Section:intro} could also be performed in a different framework, as long as they reproduce the same limitations. KITTI \cite{Geiger2012CVPR} provides a wide range of different available annotations and benchmarks for vehicle exterior applications. Closely related are the Cityscapes dataset \cite{Cordts2016Cityscapes} for different segmentation tasks, ECP \cite{braun2019eurocity} for person detection in urban traffic scenes and JTA \cite{fabbri2018learning} for pedestrian pose estimation and tracking. On the other hand, there is COCO \cite{lin2014microsoft}, a widely used benchmark for object detection, keypoint detection and panoptic and stuff segmentation as well as PASCAL VOC \cite{Everingham10}. Similarly, with Open Images \cite{OpenImages}, the largest unified dataset for image classification, object detection and instance segmentation was released. Even though these datasets contribute a wide range of images and corresponding annotations, they all have in common that their provided data has intrinsically high background and intra-class variation due to their nature for the exterior application. These datasets can be used to benchmark models for their performance and push the state-of-the-art in specific tasks, as ImageNet \cite{deng2009imagenet} did for classification. However, it is not possible to test the generalization to new environments and unseen intra-class variations for a larger range of tasks when only a limited amount of variability is available during training. In particular, those datasets cannot be used to benchmark applications for the (vehicle) interior regarding the challenges discussed in Section \ref{Section:intro}.

The annual VISDA challenge \cite{visda2017} hosts a benchmark for domain adaptation for different tasks, but it is limited to the transfer from synthetic to real data and solutions to different tasks are not comparable. It includes the Syn2Real \cite{Peng2018Syn2RealAN} dataset for classification and object detection and the transfer from GTA sceneries \cite{Richter_2016_ECCV} to Cityscapes \cite{Cordts2016Cityscapes} for segmentation. Other common datasets for domain adaptation, e.g. Office-Home \cite{venkateswara2017Deep}, DomainNet  \cite{peng2018moment} and Open MIC \cite{koniusz2018museum}, focus on a single task and/or the transfer from non-real to real environments. Some approaches combine two existing datasets to test the generalization from synthetic to real images, e.g. from synthetic traffic signs \cite{10.1007/978-3-319-02895-8_52} to real ones \cite{Stallkamp2012}.

It is believed that scene decomposition into meaningful components can improve the transfer performance on a wide range of tasks \cite{burgess2019monet}. Although datasets like CLEVR \cite{johnson2017clevr} and Objects Room \cite{burgess2019monet} exist, they are limited to toy examples and lack increased visual complexity.

Moreover, deep learning-based approaches capture too much relevance between the information contained in the background and the task the models are designed to solve \cite{tian2018eliminating}. Consequently, the aforementioned datasets all help to push the state-of-the-art for many computer vision tasks, but lack the possibility to investigate the challenges introduced in Section \ref{Section:intro}. With our SVIRO dataset and benchmark we are the first to provide the means to analyze the generalization and reliability of machine learning-based approaches for different tasks when only a limited number of variations is available during training. We thereby address an important engineering issue. Further, recent studies have shown the importance and applicability of using synthetic data for investigations in the automotive industry \cite{chen2019learning} possibly in combination with real data \cite{nowruzi2019much, tremblay2018training}.

\section{Dataset}
\label{Section:dataset}

We created a synthetic dataset to investigate and benchmark machine learning approaches for the application in the passenger compartment regarding the challenges introduced in Section \ref{Section:intro} and to overcome some of the shortcomings of common datasets as explained in Section \ref{Section:relatedwork}.

\subsection{Synthetic objects}

We used the free and open source 3D computer graphics software Blender 2.79 \cite{blender} to construct and render the synthetic 3D sceneries. We used realistic child safety seats or child restraint systems (CRS) to which we will simply refer to as child seats. For our dataset, we selected a subset of available seats on the market, from which we then created a 3D model so that it could be used in our simulation. The 3D models were generated using depth cameras (Kinect v1) and precise structured light scanners (Artec Eva).

We needed to define the reflection properties and visual colors for each 3D object in the scene, so that its perception by the camera under simulated lightning conditions could be calculated. For this, we used textures (Albedo, Normal and Roughness images) from Textures.com \cite{textures} (with permission) for all the objects in the scene. The environmental background and lightning were created by means of High Dynamic Range Images (HDRI) from HDRI Haven \cite{hdrihaven}. The human models (adults, children and babies) and their clothing (additional clothes were downloaded from the community assets \cite{makehuman}), were randomly generated by using the open source 3D graphic software MakeHuman 1.2.0 \cite{makehuman}. The 3D models of the cars were purchased from Hum3D \cite{hum3d} and everyday objects (e.g. backpacks, boxes, pillows) were downloaded from Sketchfab \cite{sketchfab}.

\subsection{Design choices}
During the data generation process we tried to simulate the conditions of a realistic application. We decided to partition the available human models, child seats and backgrounds such that one part is only used for the training images (for all the vehicles) and the other part is used for the test images. For each of the ten different vehicle passenger compartments and available child seats, we fixed the texture as if real images had been taken. Consequently, the machine learning models need to generalize to previously unknown variations of humans, child seats and environments. In this setting, we can train models in one or several car environment(s) and test them on a different one. This is an advantage compared to common domain adaptation datasets \cite{Peng2018Syn2RealAN, venkateswara2017Deep, peng2018moment, 10.1007/978-3-319-02895-8_52, Stallkamp2012} which usually focus on the transfer from synthetic to real images. Further, the photorealistic rendering and close-to-real models introduce a high visual complexity which makes them more challenging than toy examples \cite{burgess2019monet, johnson2017clevr}. The dataset has an intrinsic dominant background and texture bias: all of the images are taken in a few passenger compartments, but generalization to new, unseen, passenger compartments and child seats should be achieved. This evaluation is currently not possible by state-of-the-art datasets \cite{Cordts2016Cityscapes, braun2019eurocity, fabbri2018learning, Everingham10, Geiger2012CVPR, OpenImages, lin2014microsoft}.

\begin{figure*}
	\begin{center}
		\includegraphics[width=0.975\linewidth]{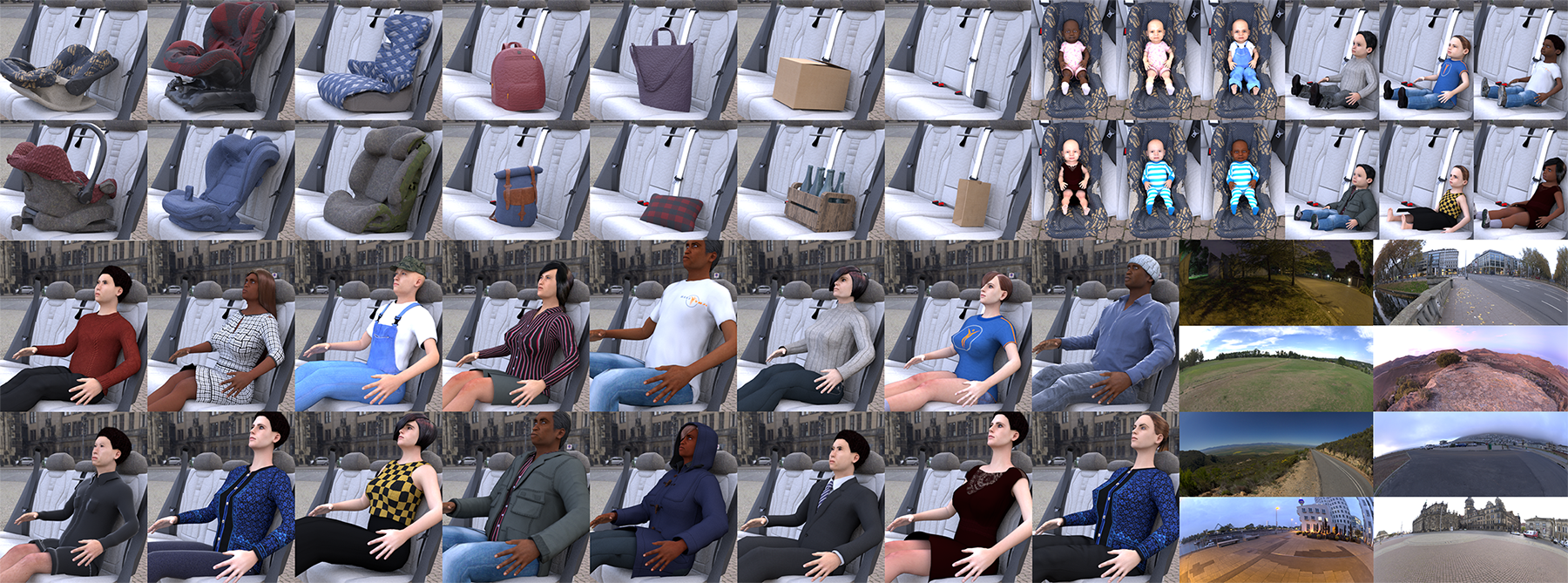}
	\end{center}
	\caption{Representative selection of the assets used for our synthetic dataset. First and third row are assets used for the training while the second and fourth are assets used for testing. Some children and adults for training and 1 environment per split are not shown. }
	\label{fig:assets}
\end{figure*}

The human models were generated randomly in MakeHuman. Their facial expression was selected to be neutral and identical. We defined a fixed set of poses for the humans represented by unit quaternions. For every human in each scenery, two poses were selected randomly and a spherical linear interpolation (Slerp) \cite{dam1998quaternions} was performed to get an intermediate pose. For each scenery, we randomly selected what kind of object is placed at each position, however, we avoided appearances of the same object for a same scenery. Child and infant seats can be empty and we decided to not allow children to be placed on the rear seat without a child seat. Infant seats were randomly rotated by $180^{\circ}$ along the z-axis and an offset from the straight ahead orientation was randomly applied to all child seats. The handle of the infant seat was selected to be up or down. Randomly selected environmental backgrounds were rotated around the vehicle to simulate arbitrary lightning conditions. We placed everyday objects onto the rear seat to make the scenery more versatile. All cameras have the same intrinsic parameters (focal length=\SI{3.4}{\milli\metre}, sensor width: \SI{8.5}{\milli\metre}, f-number$=2.5$, skew coefficient$=0$, focal length in terms of pixels: $\alpha_x=514.4208$, $\alpha_y=514.4208$, principal point: $u_0=640$, $v_0=480$), however, their pose is different in each car. Example sceneries for training and test data can be found in Figure \ref{fig:intra_car_variation} and in the supplementary material. An overview of the 3D objects are shown in Figure \ref{fig:assets}.
\begin{figure*}
	\begin{center}
		\includegraphics[width=0.975\linewidth]{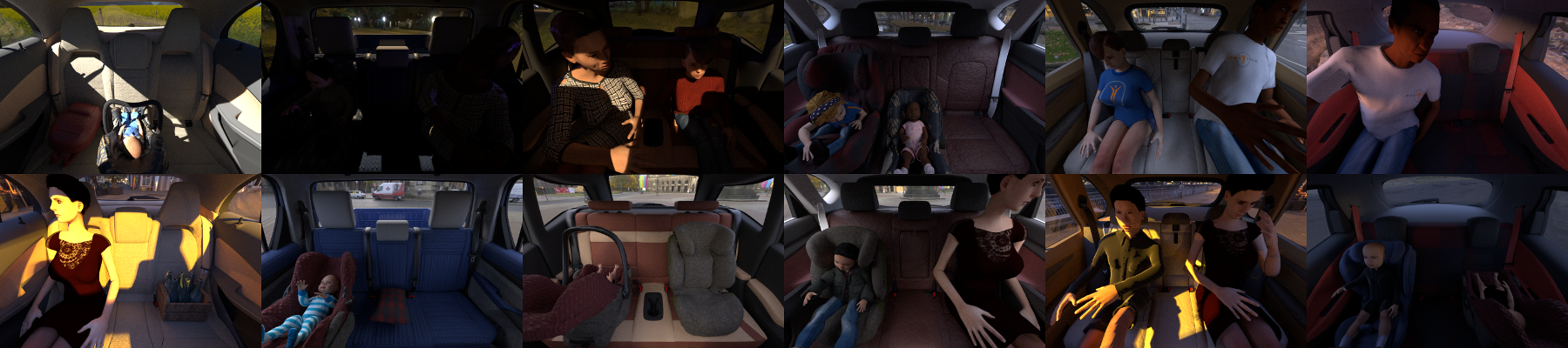}
	\end{center}
	\caption{Example sceneries for training (top) and test (bottom) splits for different cars. Each split uses different objects, seats, environments and humans. Some images appear darker, which is why (also in real applications) it is preferred to use an active infrared camera system.}
	\label{fig:intra_car_variation}
\end{figure*}

We also generated a training dataset with randomly selected (partially unrealistic) textures and backgrounds from a large pool of images. When trained on the latter, the increased variations improve the generalization for classification and semantic segmentation on the test set and to new passenger compartments, as shown in Section \ref{Section:results_classification} and \ref{Section:results_semantic}. An additional advantage of our approach is the possibility to create images under defined conditions (e.g. same scenery, but under different lightning conditions) so that additional investigations can be performed in future works. Moreover, the difficulty can be gradually increased: one can, for example, train on occupied child and infant seats only, train on infant seats with the handle down (or up) only or removing everyday object completely from training.

\subsection{Statistics}
\label{Section:statistics}
\begin{figure*}
	\begin{center}
		\begin{overpic}[width=0.975\linewidth]{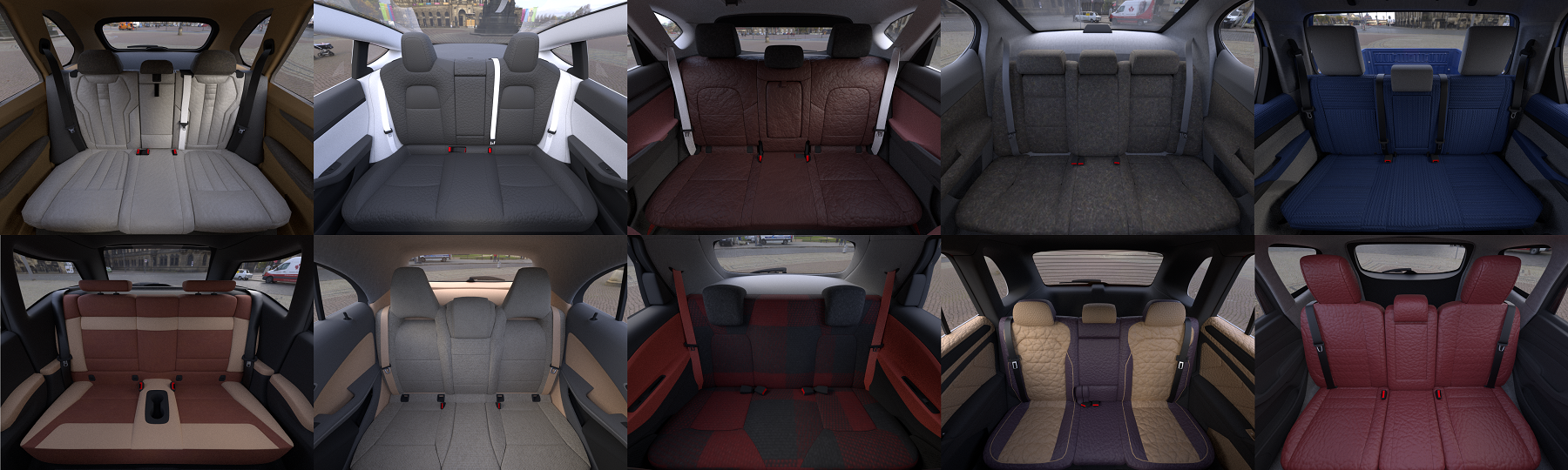}
			\put(1,16){\Large\textcolor{white}{a}}
			\put(21,16){\Large\textcolor{white}{b}}
			\put(41,16){\Large\textcolor{white}{c}}
			\put(61,16){\Large\textcolor{white}{d}}
			\put(81,16){\Large\textcolor{white}{e}}
			
			\put(1,1){\Large\textcolor{white}{f}}
			\put(21,1){\Large\textcolor{white}{g}}
			\put(41,1){\Large\textcolor{white}{h}}
			\put(61,1){\Large\textcolor{white}{i}}
			\put(81,1){\Large\textcolor{white}{j}}
		\end{overpic}
	\end{center}
	\caption{Comparison of the different vehicle interiors. a) BMW X5, b) Tesla Model 3, c) Hyundai Tucson, d) Lexus GS F, e) Toyota Hilux, f) BMW i3, g) Mercedes A-Class, h) Renault Zoe, i) VW Tiguan and j) Ford Escape. The geometry of the rear-seat, the windows, headrest and car features differ between the cars and some cars only have two seats instead of three. 
	}
	\label{fig:interior_comparison}
\end{figure*}

\begin{figure}
	\begin{center}
		\includegraphics[width=0.975\linewidth]{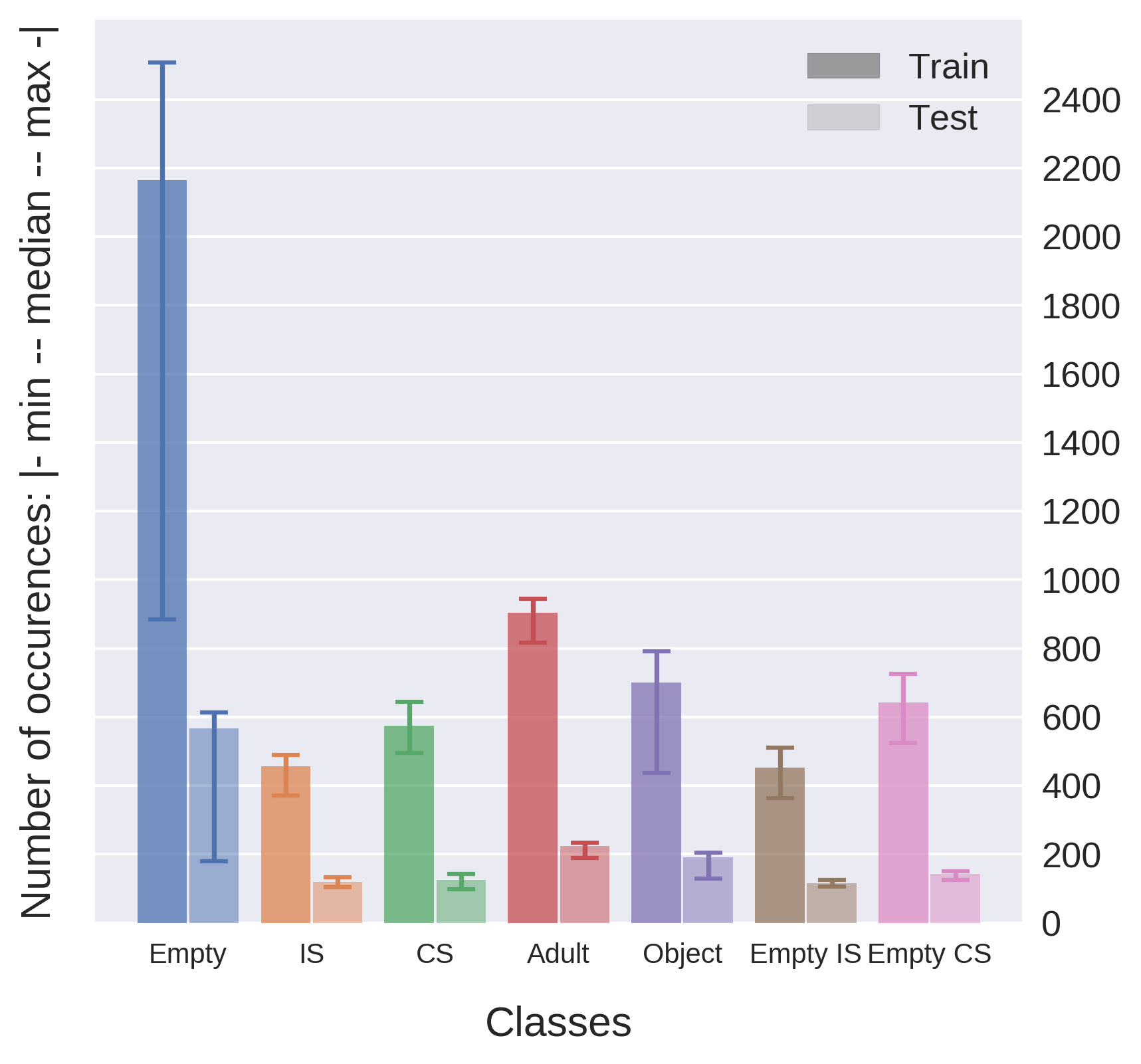}
	\end{center}
	\caption{Distribution of the different classes over the vehicles and data splits. As the images were generated randomly, the distribution is different for each split and vehicle. The bar represents the median value for a given class for a given data split over all vehicles. The error bar represents the minimum and maximum number of occurrences along the vehicles for a given split. The dark colors represent the training data and the light ones the test data. We abbreviate infant seat as IS and child seat as CS. The large difference in empty seats is due to vehicles with only two rear seats.}
	\label{fig:histogram_class_distribution}
\end{figure}

Our dataset consists of ten different vehicles: BMW X5, BMW i3, Hyundai Tucson, Tesla Model 3, Lexus GS F, Mercedes A-Class, Renault Zoe, VW Tiguan, Toyota Hilux and Ford Escape. The number of windows varies, which causes different lightning conditions, and some cars have only two rear seats instead of three. The different vehicle interiors are compared in Figure \ref{fig:interior_comparison}. We used the same people and child seats for the training set of each vehicle and the remaining ones for the test sets. This results in two child seats and one infant seat per data split. We did the same for the background: five were selected for the training and five different ones for the test set. For the everyday objects, we used two bags, a card-box and a cup for the training dataset and a different bag, a paper-bag, pillows and a box of bottles for the test set. The number of people and the distribution of the gender, age and ethnicity for the training and test set can be found in Table \ref{table:statistics}. The number of images generated for each vehicle and each training and test set are identical. In total, this results in $20000$ training and $5000$ test sceneries. The distribution of the different classes across the vehicles and data splits is summarized in Figure \ref{fig:histogram_class_distribution}. The number and constellation of appearances varies between the vehicles, because all the sceneries were generated randomly.
\begin{table}
	\begin{center}
		\begin{tabular}{|p{1.35cm}|P{0.625cm}|P{0.625cm}|P{0.625cm}|P{0.625cm}|P{0.625cm}|P{0.625cm}|}
			\cline{2-7}
			\multicolumn{1}{c|}{} & \multicolumn{3}{c|}{Train} & \multicolumn{3}{c|}{Test}\tstrut \\
			\cline{2-7}
			\multicolumn{1}{c|}{} & \small Adult & \small Child & \small Baby & \small Adult & \small Child & \small Baby\tstrut \\
			\cline{1-7}
			African & 5 & 2 & 1 & 2 & 1 & 1\tstrut \\
			Asian & 5 & 2 & 1 & 2 & 2 & 1\tstrut \\
			Caucasian & 4 & 2 & 1 & 4 & 1 & 1\tstrut\\
			\cline{1-7}
			Female  & 9 & 3 & - & 5 & 2 & -\tstrut \\
			Male  & 5 & 3 & - & 3 & 2 & -\tstrut \\
			\cline{1-7}
			Total & 14 & 6 & 3 & 8 & 4 & 3\tstrut \\
			\hline\hline
			Per Car & \multicolumn{3}{c|}{2000} & \multicolumn{3}{c|}{500}\tstrut \\
			\hline
		\end{tabular}
	\end{center}
	\caption{Number of people and distribution of gender, age and ethnicity for the training and test dataset. The same people were used for the training and test set for all the vehicles, respectively, and the same number of images were generated for each car.}
	\label{table:statistics}
\end{table}

\subsection{Rendering}

\begin{figure}
	\begin{center}
		\begin{overpic}[width=0.325\linewidth]{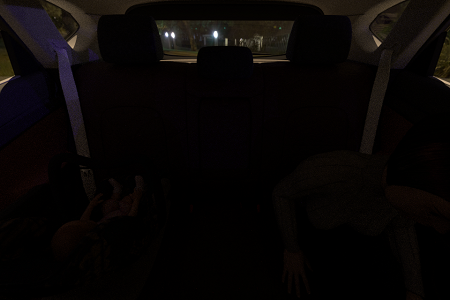}
			\put(4,4.7){\Large\textcolor{white}{a}}
		\end{overpic}
		\begin{overpic}[width=0.325\linewidth]{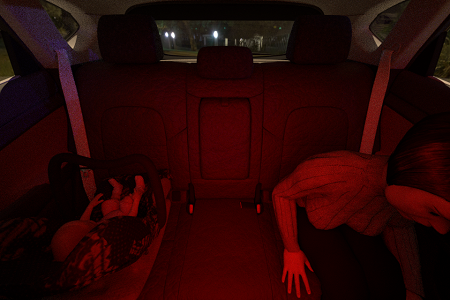}
			\put(4,4.7){\Large\textcolor{white}{b}}
		\end{overpic}
		\begin{overpic}[width=0.325\linewidth]{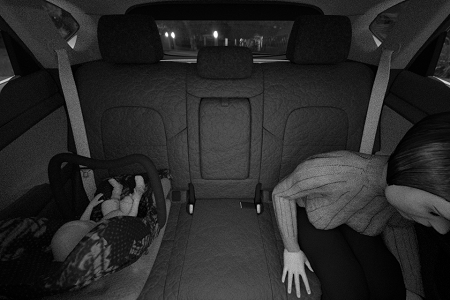}
			\put(4,4.7){\Large\textcolor{white}{c}}
		\end{overpic}
	\end{center}
	\caption{Comparison between a standard RGB image and our simple approach to imitate an active infrared camera system for a dark scenery. a) Standard RGB image in environmental lightning. b) RGB image of the scenery illuminated by an active red light. c) Red channel only of the RGB image of the illuminated scenery (used as infrared imitation in SVIRO).}
	\label{fig:compare_ir}
\end{figure}

The synthetic images were generated using Blender, its Python API and the Cycles renderer. As many applications in the passenger compartment require an active infrared camera system to work in the dark, we decided to imitate such a system by means of a simple approach: We placed an active red lamp (R=100\%, G=0\%, B=0\%) next to the camera inside of the car illuminating the rear seat, but overlapping with the illumination from the HDR background image. We then took the red channel only from the resulting rendered RGB image. We will refer to these images as grayscale images. This is, however, not a physically accurate simulation of a real active infrared camera system. The simulation of the latter is not trivial, as the perception in the infrared domain not only depends on the object's material properties, but also on the wavelength which is used \cite{piazena2017spectral}. We argue that this is of minor importance, because SVIRO is intended to investigate the general applicability of possible machine learning methods. Our approach helps to become less dependent on the environmental lightning and to facilitate the tasks: see Figure \ref{fig:compare_ir} for a comparison between a standard RGB image and our grayscale image for a dark scenery, where a lot of information would be lost. More comparisons are available in the supplementary material. Moreover, we report in Section \ref{Section:real_application} and Figure \ref{fig:compare_real} the evaluation of a model trained on SVIRO on real infrared images and show that it behaves similarly on real data.

\subsection{Ground truth}
\label{Section:ground_truth}

\begin{figure}
	\begin{center}
		\includegraphics[width=\linewidth]{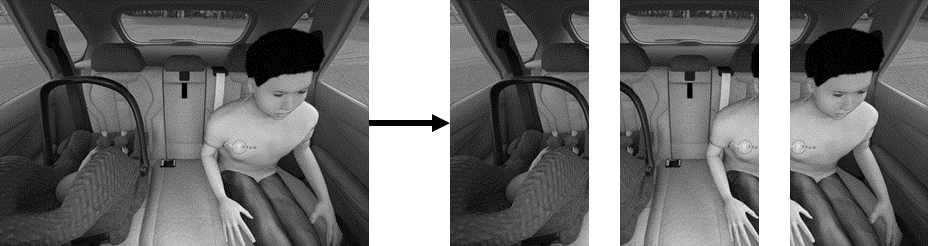}
	\end{center}
	\caption{We split each image into three rectangles to use them for classification. The contents of the rectangles overlap slightly, because objects are not limited to their seat position.}
	\label{fig:splitted}
\end{figure}

For each scenery we provide a set of images and ground truth data: 1) An RGB image of the scenery without an active red lamp next to the camera, e.g. Figure \ref{fig:intra_car_variation}, 2) a grayscale image (red channel only) of the rendered RGB image using an active red lamp next to the camera, e.g. Figure \ref{fig:example_real_orss} (b), 3) an instance segmentation map, where each object is color-coded depending on its position and class, e.g. Figure \ref{fig:example_real_orss} (c), 4) Bounding boxes for all the elements in the scenery, 5) Keypoints for all the human poses in the scenery, e.g. Figure \ref{fig:example_real_orss} (a), 6) a depth map of the scenery, e.g. Figure \ref{fig:example_real_orss} (d).
For classification, we split the images (RGB, grayscale, depth) into three rectangles (one for each seat position) with slight overlap between them. See Figure \ref{fig:splitted} for an illustration. If a car has only two seats, then we exclude the middle rectangle. Note that objects from neighbouring seats are overlapping to the neighbouring rectangle, which makes classification more difficult. However, this is necessary as people can lean over to the neighbouring seat. Both semantic segmentation and instance segmentation can be performed using the provided segmentation masks. Children on a child seat, as well as babies in an infant seat, are treated as two separate instances. We save the human poses by using keypoints, as used by the COCO dataset \cite{lin2014microsoft}, but our skeleton is defined using partially different joints. The visibility of the keypoints are set to zero if the keypoint is outside the image, to one if it is occluded by an object or neighbouring human and set to two if it is visible or occluded by the person itself. Keypoints are provided for the babies as well. For each scenery, we provide a .json file containing the 2D pixel coordinates of the keypoints of all people together with the visibility flag, the bone names and their seat position. All the images are provided in .png format. The depth maps are saved in millimetres and as 16-bit .png images. The bounding boxes are given in the format $[$class, $x_1$, $y_1$, $x_2$, $y_2]$, where ($x_1$, $y_1$) is the upper left corner and ($x_2$, $y_2$) the lower right corner of the bounding box (coordinates start in the upper left image corner). For classification, the labels are as follows: 0=empty seat, 1=infant in infant seat, 2=child on child seat, 3=adult passenger, 4=everyday object, 5=infant seat without baby, 6=child seat without child. For segmentation and object detection, the labels are: 0=background, 1=infant seat, 2=child seat, 3=person and 4=everyday object. We did not fasten the seat-belt for our models and let them un-attached in all our sceneries.

\section{Baseline evaluation}
\label{Section:results}

In this baseline evaluation, we will show that SVIRO provides the means to analyze the performance of common machine learning methods under new conditions. We will test some widely used models and approaches for their robustness and reliability, when trained on limited number of variations only. Specifically, we will show that state-of-the-art models cannot generalize well to new environments and textures when trained on the previously discussed challenging, but realistic, conditions. For this evaluation, we limited ourselves to training on the X5 and testing on the Tucson (three seats) and i3 (two seats). For all tasks, we considered two training data versions (for which we used the exact same hyper-parameters): 1) the standard X5 training data with fixed textures and backgrounds (F), 2) half of the standard X5 training data is replaced by randomly textured X5 training data with random backgrounds (F\&R).

We used the grayscale images (infrared imitation) for all the evaluations. For the deep learning-based approaches, we used the pre-defined models implemented in PyTorch 1.2 and Torchvision 0.4.0. For classification, we used pre-trained models on ImageNet. For semantic and instance segmentation, the models were pre-trained on COCO train 2017. The pre-trained models were fine-tuned on the X5 only and then evaluated on the test sets of all three cars. Using this approach, we could test the generalization capacities on two difficulty levels. The training dataset was partitioned randomly according to a 75:25 split for training and evaluation, where the latter was used to perform early stopping when fine-tuning the models. As we consider our F\&R dataset as data augmentation, the only additional data augmentation performed was a random horizontal flip.

\subsection{Classification}
\label{Section:results_classification}
\begin{figure*}
	\begin{center}
		\includegraphics[width=\linewidth]{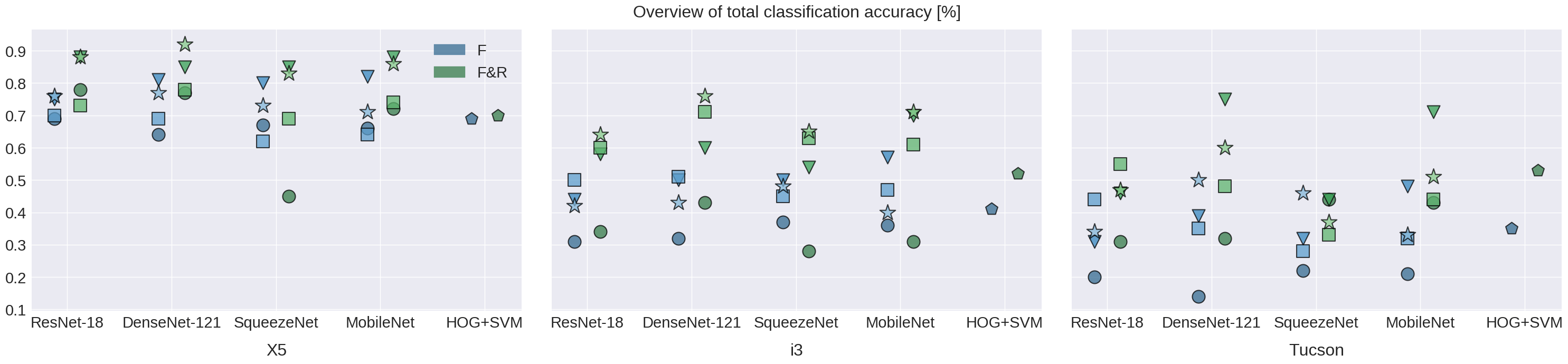}
	\end{center}
	\caption{Comparison of different classification results. We trained several models from scratch ($\medblacksquare$) or fine-tuned pre-trained models, where all the weights ($\medblacktriangledown$), the last block ($\medblackcircle$) or the last layer ($\medblackstar$) were trainable. Further, we trained a SVM using HOG features. We used the standard X5 training data (F, in blue) or replaced half of it with the randomized data (F\&R, in green). After training, we retained models with the best total accuracy on the X5 test data and evaluate them on the i3 and Tucson test data. The models have difficulties to generalize to the test data and perform even worse in unknown vehicles, but including the randomized data helps to generalize to unseen objects.}
	\label{fig:classification_result}
\end{figure*}
As introduced in Section \ref{Section:ground_truth}, we used the rectangular graycale images for classification with seven different classes. One could decide to classify a seat with an everyday object (and an empty infant/child seat) as empty as well. We trained a single classifier for the three seats, but other setups are possible as well, e.g. train one classifier for each seat. In the following, we will report results on different deep learning models, as they are commonly used for visual classification problems. These results will be compared to a traditional method using a support vector machine (SVM) and handcrafted features. We will show that both methods suffer from the same problems and including the randomized F\&R dataset overall improves the results.

\subsubsection{CNN}

We used the ResNet \cite{he2016deep}, DenseNet \cite{huang2017densely}, SqueezeNet V1.1 \cite{iandola2016squeezenet} and MobileNet V2 \cite{sandler2018mobilenetv2} architectures and considered four different training approaches: 1) Training from scratch, 2) only fine-tuning the last fully connected layer, 3) additionally fine-tuning the last residual block, 4) allowing all weights to be trainable. We tried different combinations of weight decay, weighted costs and imbalanced sampling and report results for the best models only. In Figure \ref{fig:classification_result}, we compare the results across the different models and training approaches and compare them to the SVM. The deep learning-based approaches have problems to generalize to the test set, especially for new cars. The randomized backgrounds and textures help to improve the accuracy on the same car, which gives hint that models trained on the (F) dataset mostly use the texture as a classification criterion. However, the models can still not generalize well to new vehicle interiors, probably because of the different interior structures (see Figure \ref{fig:interior_comparison}). An exhaustive comparison between the different training approaches and hyper-parameters is available in our supplementary material.

\subsubsection{HOG and SVM}

For comparison, we also wanted to test at least one traditional machine learning-based approach for the classification task. To this end, we computed the histogram of oriented gradients (HOG) features of all the training images, and their horizontally flipped versions for data augmentation. These features were then used to train a SVM, using the "one vs. rest" approach and balanced class weights. We performed a grid search on different kernels (linear, polynomial and radial basis) and their hyper-parameters and used a 5-fold cross validation for hyper-parameter selection. We used scikit-learn 0.21.2 for the training and scikit-image 0.15.0 for the feature generation. The results for the best hyper-parameters are reported in Figure \ref{fig:classification_result}. Overall, the traditional approach has similar problems as the deep learning approach when the standard X5 data is used, and can sometimes even generalize better. However, it cannot exploit the additional information when random textures and backgrounds are included in the training.

Our dataset shows that traditional and deep learning approaches, although commonly used in practice, drastically decrease classification performance when trained in a setting with limited variations without taking additional precautions. No reliability can be guaranteed and both presented approaches do not fully grasp the underlying task, although the environment and the objects are similar. Including randomized images increases the performance, but to be applicable in real world applications further (theoretical) improvements need to be investigated and developed.

\subsection{Semantic segmentation}
\label{Section:results_semantic}

\begin{figure}
	\begin{center}
		\includegraphics[width=\linewidth]{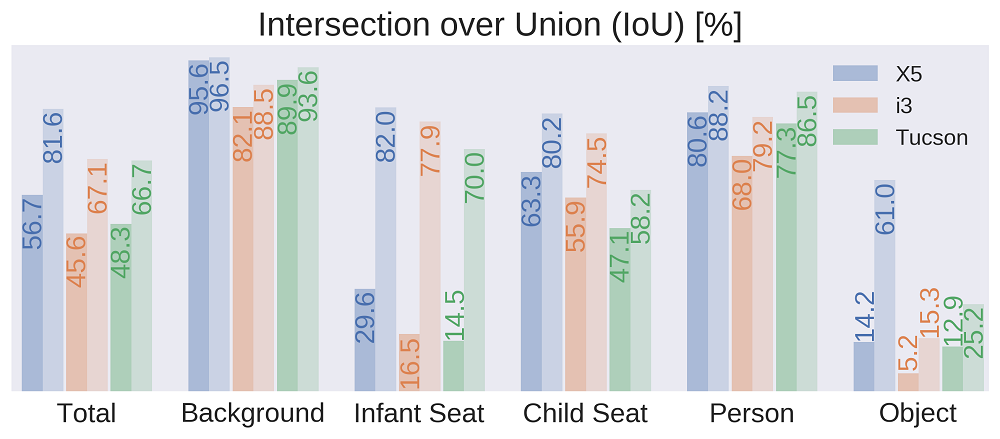}
	\end{center}
	\caption{Mean intersection over union (IoU), in percent, for semantic segmentation for a fine-tuned pre-trained FCN. The dark colour represents the model performance when trained on the standard X5 training dataset (F) and the lighter colour when we included the random X5 data (F\&R). The models were evaluated on the test dataset for the X5, Tucson and i3. Using the randomized version largely improves the generalization capacities of the model, especially for identifying infant seats and child seats.
	}
	\label{fig:segmentation_result}
\end{figure}

It could be beneficial to take spatial information into account to improve the transfer to new instances and environments. Further, the model might consider overlapping objects from neighbouring seats more efficiently when the entire scene is used. To this end, we evaluated semantic segmentation and considered the five classes as introduced in Section \ref{Section:ground_truth}. The model should separate the child from the child seat and the baby from the infant seat and classify them as a people. We fine-tuned all layers of a Fully Convolutional Network (FCN) with a ResNet-101 backbone and report the results in Figure \ref{fig:segmentation_result}. As for the classification results of the previous section, the model's performance decreases drastically on the child and infant seats on the test set for the same car and it performs even worse in previously unknown cars. Using the F\&R training data, the generalization performance largely increased, although the geometry of the child seats of the test sets was never observed during training. It seems that the texture has a larger influence on the performance of classification and semantic segmentation models than the geometry. This observation seems to be in line with recent results by Geirhos \etal \cite{geirhos2018imagenet}. However, using SVIRO, we can additionally show that the model cannot perform as good on new environments, even though the textures are randomized and the objects of the different test sets are the same. 

\section{Comparison with real images}
\label{Section:real_application}
\begin{figure}
	\begin{center}
		\begin{overpic}[width=0.48\linewidth]{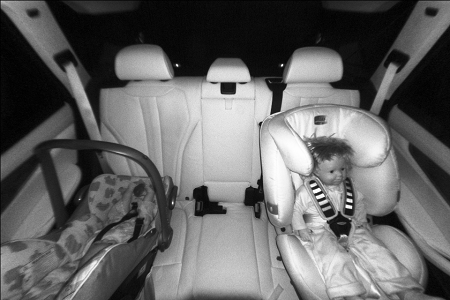}
			\put(39,55){\Large\textcolor{white}{Real}}
		\end{overpic}
		\begin{overpic}[width=0.48\linewidth]{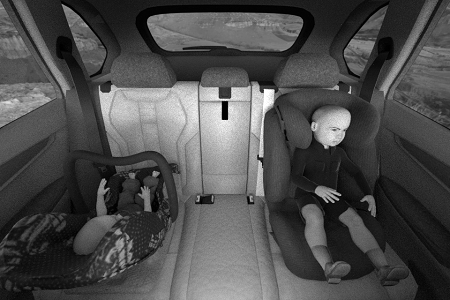}
			\put(32,55){\Large\textcolor{white}{SVIRO}}
		\end{overpic}
		\vskip 0.1 \baselineskip
		\includegraphics[width=0.48\linewidth]{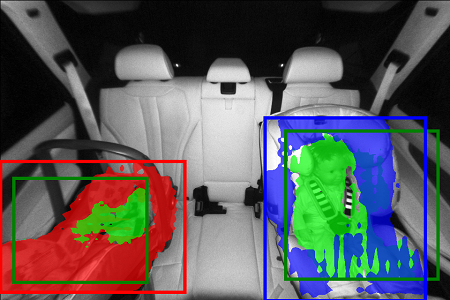}
		\includegraphics[width=0.48\linewidth]{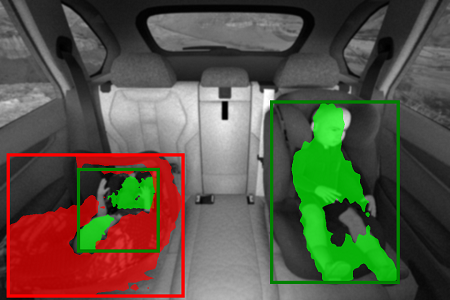}
		\vskip 0.1 \baselineskip
		\includegraphics[width=0.48\linewidth]{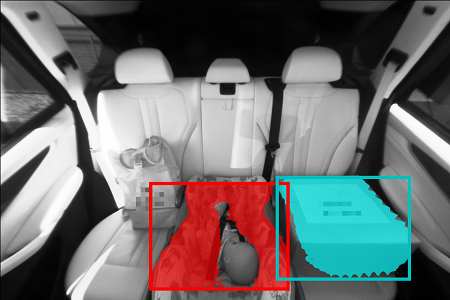}
		\includegraphics[width=0.48\linewidth]{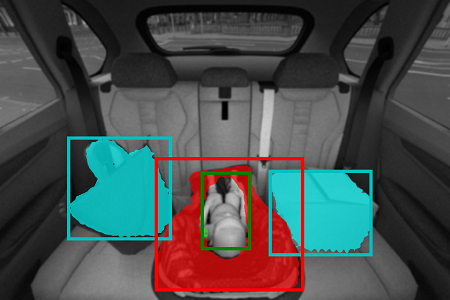}
		\vskip 0.1 \baselineskip
		\includegraphics[width=0.48\linewidth]{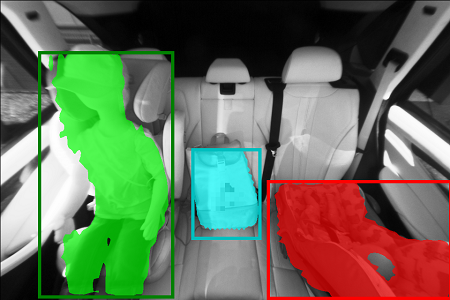}
		\includegraphics[width=0.48\linewidth]{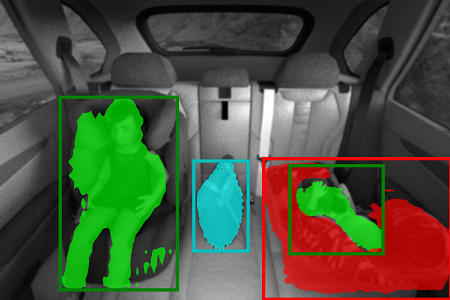}
		\vskip 0.1 \baselineskip
		\includegraphics[width=0.48\linewidth]{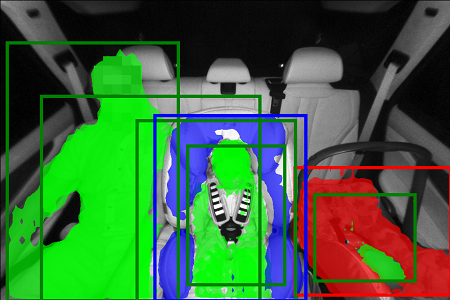}
		\includegraphics[width=0.48\linewidth]{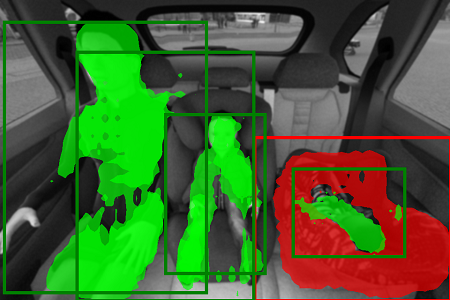}
	\end{center}
	\caption{We acquired real active infrared images (first column) in an X5 and reproduced the same sceneries in Blender (second column). The first row compares real and synthetic images. The remaining rows compare instance segmentation mask predictions. The model performs similarly on both setups and the similar child seat is detected in the real images, but not in the synthetic ones. 
	}
	\label{fig:compare_real}
\end{figure}
We tested the transferability of a model trained on SVIRO to real infrared images and report results on instance segmentation to illustrate this. We fine-tuned all layers of a pre-trained Mask R-CNN model with a ResNet-50 backbone and considered the same classes as for semantic segmentation. The synthetic images were blurred to be closer to real infrared images. We combined the training images of the i3, Tucson and Model 3 and compare results on synthetic and real images in the X5 in Figure \ref{fig:compare_real}. More evaluations on real images are available in the supplementary material. Only bounding boxes and masks with a confidence of at least 0.5 are plotted. The model performs similarly across real and synthetic images and sometimes fails to detect objects. This is expected as the model has only seen a limited amount of variation. However, the similar child seat is detected in the real images, but not in the synthetic ones. We believe that investigations on SVIRO are transferable to real applications as the resulting model behaves similarly on real and synthetic images. Additional realistic effects could be applied to close the synthetic gap even further \cite{ley2016syb3r}.

\section{Conclusion}

We release SVIRO, a synthetic dataset for sceneries in the passenger compartment of ten different vehicles. Our benchmark addresses real-world engineering obstacles regarding the robustness and generalization of machine learning models. Using SVIRO, we showed in our baseline evaluation that common machine learning models, when trained on limited amount of variability, decrease in performance for solving the same task in a new vehicle interior. Models cannot generalize well to new intra-class variations, even in the car they were trained on. We believe that other research directions, e.g. (disentangled) latent space representation, scene decomposition, domain adaptation and uncertainty estimation, can benefit from our dataset.

\noindent\textbf{Acknowledgement:} The first author is supported by the Luxembourg National Research Fund (FNR) under the grant number 13043281. This work was partially funded by the MECO project ”Artificial Intelligence for Safety Critical Complex Systems” and the  European Union’s Horizon 2020 Program in the project VIZTA (826600).

{\small
	\bibliographystyle{ieeetr}
	\bibliography{bib}
}

\end{document}